\documentclass{article}
\usepackage{spconf,amsmath,graphicx}

\usepackage{graphicx}
\usepackage{algorithm}
\usepackage{algorithmic}
\usepackage{pifont}
\usepackage[most]{tcolorbox}
\usepackage{booktabs}
\usepackage{amsfonts}
\usepackage{url}

\usepackage{listings}

\title{Scene-Action Prompt Fusion for Coherent Text-to-Video Storytelling}

\name{
    Taewon Kang$^{1}$, Divya Kothandaraman$^{2}$, Ming C. Lin$^{1}$
}
\address {$^{1}$University of Maryland at College Park, United States, $^{2}$Dolby Laboratories, United States \\
    {\large \texttt{taewon@umd.edu, divya.kothandaraman@dolby.com, lin@umd.edu}} \\
}

\begin{document}

\maketitle

\vspace*{-4.5em}
\begin{abstract}
Generating coherent long-form video sequences from discrete text prompts remains challenging due to difficulties in maintaining temporal coherence, semantic consistency, and scene–action continuity across segments. We propose a novel storytelling framework that integrates scene and action prompts through dynamics-inspired prompt mixing. Our approach combines three key components: (i) a bidirectional time-weighted latent blending strategy that enforces temporal consistency between consecutive video segments, (ii) a dynamics-informed prompt weighting (DIPW) mechanism that adaptively balances scene and action prompts at each diffusion timestep based on CLIP-based alignment, narrative progression, and temporal smoothness, and (iii) a semantic action representation that encodes high-level action semantics to modulate transitions according to action similarity. Latent-space blending preserves spatial coherence within scenes, while time-weighted blending introduces bidirectional temporal constraints to prevent abrupt transitions. Together, these components enable fluid and coherent video narratives that faithfully reflect both scene context and action dynamics. Extensive experiments demonstrate that our method significantly outperforms baselines, producing temporally consistent and visually compelling long-form videos without any additional training, thereby bridging the gap between short clips and extended text-driven video storytelling.
\end{abstract}


\begin{figure*}[htb!]
\vspace*{-7em}
\begin{center}
\includegraphics[width=0.9\linewidth]{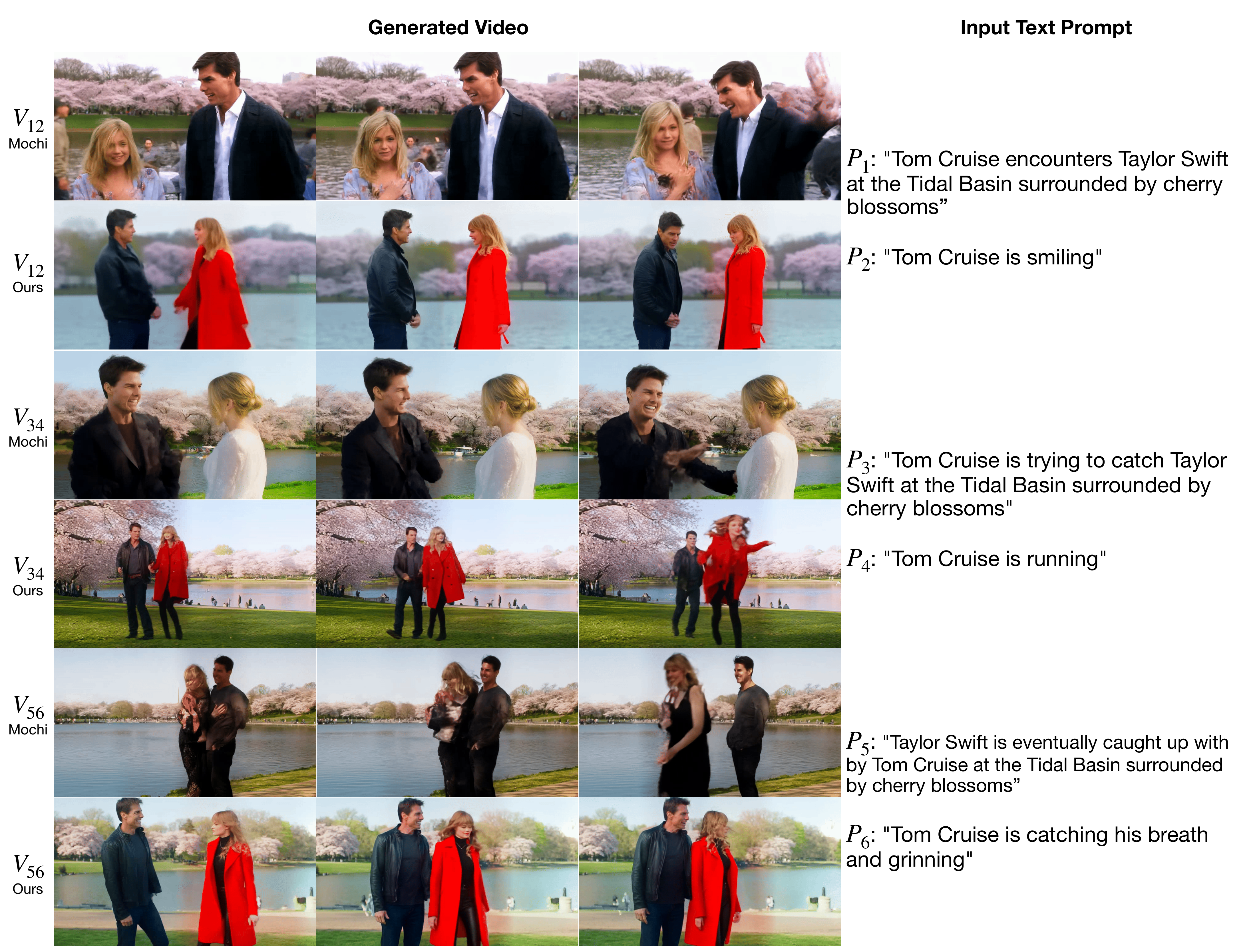}
\end{center}
\vspace*{-2em}
  \caption{\textbf{Example of story generation using our proposed method} Each row represents a video segment $V_{12}$, $V_{34}$, $V_{56}$, where the scene and motion evolve according to the specified textual prompts. 
  The first prompt ($P_1$) describes a detailed action (e.g., "Tom Cruise is trying to catch Taylor Swift at the Tidal Basin surrounded by cherry blossoms"), with the second prompt ($P_2$) follows (e.g., "Tom Cruise is running"). Our approach ensures smooth transitions between these states using Time-weighted Blending (TWB), Dynamics-Informed Prompt Weighting (DIPW), and Semantic Action Representation (SAR), preserving consistency across entire story videos. Our method (Bottom, Even Row) provides much more consistent scenes compared to the text-to-video generation model, Mochi baseline (Top, Odd Row), enabling smooth video storytelling.} 
\label{teaser_figure}
\vspace*{-1em}
\end{figure*}

\vspace*{-1em}
\section{Introduction}
\vspace*{-1em}

Generating coherent long-form videos from text prompts is a fundamental challenge in content creation~\cite{li2019storygan, maharana2021improving, zhuang2024vlogger, he2023animate, he2024dreamstory, zhao2024moviedreamer, wang2024dreamrunner, zheng2024temporalstory, li2024vstar}. 
Although recent diffusion-based models have achieved impressive results in short video synthesis, extending them to long-form storytelling remains difficult. When generation is prolonged beyond a few seconds, temporal inconsistencies, semantic drift, and disrupted scene–action continuity frequently emerge, degrading narrative flow~\cite{mochi, yin2023nuwa, qiu2023freenoise, wang2023gen, oh2024mevg, zheng2024temporalstory, villegas2022phenaki}. A common strategy for text-based video storytelling is to generate a sequence of short clips, each guided by a distinct prompt describing a different story segment~\cite{zhuang2024vlogger, he2023animate, he2024dreamstory, zhao2024moviedreamer, wang2024dreamrunner}. However, such approaches struggle to maintain temporal coherence across clips while preserving semantic meaning and action continuity. In practice, this often leads to abrupt transitions, inconsistent character appearance or motion, and fragmented narratives. Prior work such as DreamRunner~\cite{wang2024dreamrunner} incorporates retrieval-augmented adaptation and spatio-temporal attention to improve motion binding, yet still fails to ensure consistent scene-to-scene transitions in long-form videos.

To address these challenges, we propose a novel storytelling framework that enables seamless long-form video generation from discrete text prompts by integrating scene and action descriptions through dynamics-inspired control. Our method introduces three key components.  \textbf{First}, we adopt a dual-prompt formulation that explicitly separates scene context from action specification and propose \textit{Dynamics-Informed Prompt Weighting (DIPW)}, which adaptively balances the influence of scene and action prompts throughout the diffusion denoising process. This mechanism allows the model to capture both thematic richness and fine-grained action dynamics within each segment.  \textbf{Second}, we introduce \textit{Time-Weighted Blending (TWB)} with bidirectional constraints, which softly integrates latent states from neighboring segments to enforce smooth temporal transitions and prevent abrupt appearance or motion changes.  \textbf{Third}, we propose a \textit{Structured Semantic Action Representation (SAR)} that encodes high-level action semantics using a pretrained text encoder (e.g., CLIP) and modulates blending strength based on action similarity, preserving logical action progression across segments.

By unifying these dynamics-inspired components, our framework generates extended video sequences that maintain temporal and spatial coherence while exhibiting logically progressive storytelling. Extensive qualitative and quantitative evaluations demonstrate that our approach significantly outperforms prior methods, achieving the highest CLIP-add (+3.91\% vs. Mochi, +5.74\% vs. Vlogger), CLIP-combined (+3.49\%, +7.10\%), and DINO (+1.75\%, +4.78\%) scores. These results show that our method effectively bridges the gap between isolated short clips and coherent long-form video storytelling, establishing a practical and training-free framework for text-driven video synthesis.

\vspace*{-0.5em}
\section{Related Works}
\vspace*{-1em}

\subsection{Storytelling Video Generation}
\vspace*{-0.75em}
Several works leverage large language models (LLMs) to decompose scripts into sequential conditions, generating videos segment by segment. VideoDirectorGPT~\cite{lin2023videodirectorgpt} and Vlogger~\cite{zhuang2024vlogger} follow this paradigm, while Animate-A-Story~\cite{he2023animate} improves motion control by retrieving depth-conditioned reference videos. More recent methods such as DreamStory~\cite{he2024dreamstory} and MovieDreamer~\cite{zhao2024moviedreamer} generate keyframes using text-to-image models and animate them with image-to-video pipelines to enhance temporal coherence. Customization and adaptation techniques have also been explored to improve character consistency across scenes~\cite{wang2024dreamrunner,he2024dreamstory,zhao2024moviedreamer}. Despite these advances, most existing approaches rely on keyframe-based animation, motion priors, or retrieval-augmented adaptation, which limits their ability to produce smooth, dynamic motion transitions in long-form videos. DreamRunner~\cite{wang2024dreamrunner} introduces retrieval-augmented motion adaptation and spatio-temporal 3D attention, but depends on motion prior training and large-scale LLMs, resulting in high computational cost and limited scalability, while still failing to ensure consistent scene-to-scene transitions. StoryDiffusion~\cite{zhou2025storydiffusion} improves visual consistency via modified self-attention and a semantic motion predictor, yet requires additional training and primarily targets image-to-video or comic generation rather than story-driven synthesis from pure text prompts. In contrast to prior work, our approach enables controlled motion evolution without additional training or external motion priors. By explicitly decomposing prompts into \textbf{scene descriptions and action commands}, our framework performs prompt mixing that jointly captures scene evolution and action continuity. This design reflects the nature of storytelling, where narratives are shaped not only by static scenes but by interactions and actions unfolding over time, allowing our method to model both background progression and foreground dynamics within a unified diffusion process.
\vspace*{-1.75em}
\subsection{Text-to-Video Diffusion}
\vspace*{-0.75em} 
Representative models such as Mochi~\cite{mochi} employ diffusion with multi-stage conditioning to improve motion fidelity and temporal consistency, while CogVideoX~\cite{yang2024cogvideox}, building on CogVideo~\cite{hong2022cogvideo}, adopts transformer-based architectures to generate high-resolution videos with long-range coherence. However, these methods are primarily designed around \emph{single descriptive prompts} and focus on short or medium-length video generation. As a result, they lack explicit mechanisms for maintaining narrative structure, action continuity, and temporal coherence across multiple segments required for long-form storytelling. Our work complements these advances by introducing a training-free, prompt-driven framework that directly addresses multi-segment narrative consistency within pretrained text-to-video diffusion models.

\begin{figure*}[htb!]
\begin{center}
\vspace*{-7em}
\includegraphics[width=0.95\linewidth]{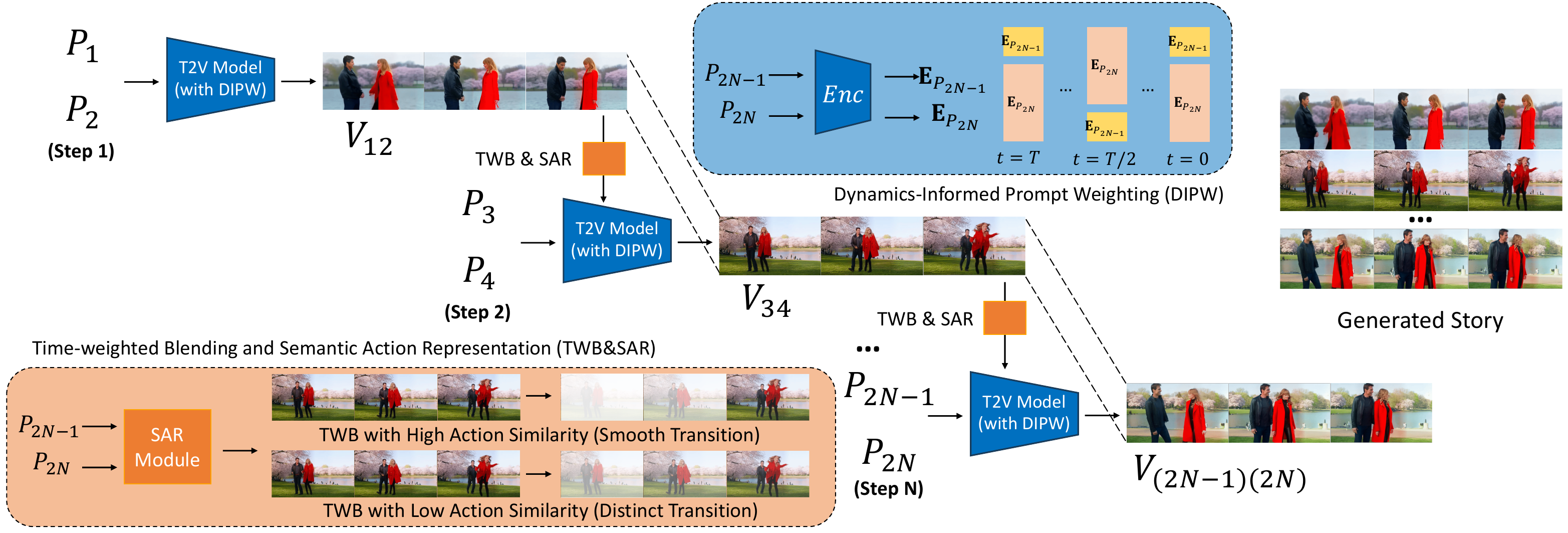}
\end{center}
\vspace*{-1em}
    \caption{\textbf{Overview of our proposed storytelling framework.} Each video segment \(V_{2N-1,2N}\) is generated from a pair of prompts \((P_{2N-1}, P_{2N})\), ensuring coherent short-form synthesis. These segments are then sequentially combined to form a complete story. We incorporate Dynamics-Informed Prompt Weighting (DIPW), Time-Weighted Blending (TWB), and Semantic Action Representation (SAR) to maintain temporal coherence and logical action continuity across segments. TWB ensures smooth transitions by dynamically adjusting blending weights based on prior frames, while SAR refines transitions based on action similarity. The framework builds upon the Mochi model, with DIPW adaptively balancing scene and action prompts to ensure structured motion evolution. Transitions between segments adaptively vary, with high action similarity producing smooth transitions and low action similarity allowing distinct but logical shifts in motion.}
\label{method_figure}
\vspace*{-1em}
\end{figure*}

\vspace*{-0.5em}
\section{Method}
\vspace*{-0.5em}

\subsection{Problem Statement}
Our goal is to generate coherent long-form video sequences from text using pretrained text-to-video models, without additional training or finetuning.
The input consists of paired text prompts $(P_1, P_2), (P_3, P_4), \ldots, (P_{2N-1}, P_{2N})$, where each pair defines a story segment: $P_{2N-1}$ describes the scene context and $P_{2N}$ specifies character actions.
The output is a sequence of video segments $\{V_{12}, V_{34}, \ldots, V_{(2N-1)(2N)}\}$, where each segment is generated independently but temporally linked to maintain narrative and motion continuity.

\subsection{Overview}
Pretrained text-to-video models perform well on short clips but struggle with long-form storytelling due to prompt interference, abrupt motion transitions, and semantic drift.
We address these issues with three components:
\textbf{(i) Dynamics-Informed Prompt Weighting (DIPW)}, which adaptively balances scene and action prompts during denoising,
\textbf{(ii) Time-Weighted Blending (TWB)}, which softly propagates latent information across segments, and
\textbf{(iii) Semantic Action Representation (SAR)}, which modulates blending strength based on action similarity.
Together, these components enable training-free long-form video generation with improved temporal coherence.

\vspace*{-0.8em}
\subsection{Dynamics-Informed Prompt Weighting (DIPW)}
DIPW dynamically adjusts the relative influence of scene and action prompts within each diffusion trajectory.
Rather than using fixed prompt weights, we define a \emph{dynamic prompting window} implicitly over diffusion timesteps, where early timesteps emphasize scene context and later timesteps emphasize action execution. At each diffusion step $i \in [1,T]$, we compute weights $\alpha_{2N-1}^i$ and $\alpha_{2N}^i$ for prompts $P_{2N-1}$ and $P_{2N}$.
These weights are determined by three complementary signals:
\begin{itemize}
    \item $\text{sim}_{2N-1}^i$, $\text{sim}_{2N}^i$: CLIP similarity between the current decoded frame and each prompt, capturing instantaneous semantic alignment;
    \item $\text{prev\_sim}_{2N-1}^i$, $\text{prev\_sim}_{2N}^i$: cosine similarity between the current prompt embedding and the combined conditioning embedding from timestep $i-1$, encouraging temporal smoothness;
    \item $\text{prior}_{2N-1}^i = 1 - \frac{i}{T}$, $\text{prior}_{2N}^i = \frac{i}{T}$: a linear progression prior that encodes the expected narrative shift from scene setup to action execution.
\end{itemize}

Both prompts are scored using the same functional form, applied independently to $P_{2N-1}$ and $P_{2N}$:
\begin{align}
s_{2N-1}^i &= \lambda_1 \text{sim}_{2N-1}^i + \lambda_2 \text{prev\_sim}_{2N-1}^i + \lambda_3 \text{prior}_{2N-1}^i, \\
s_{2N}^i &= \lambda_1 \text{sim}_{2N}^i + \lambda_2 \text{prev\_sim}_{2N}^i + \lambda_3 \text{prior}_{2N}^i .
\end{align}

The use of CLIP similarity for current-frame alignment and cosine similarity for embedding continuity reflects their distinct roles: CLIP captures multimodal semantic grounding, while cosine similarity stabilizes prompt interpolation across timesteps. Although the progression prior is uniform across segments, it is required to disambiguate prompt dominance when semantic similarities are comparable. Scores are normalized via a temperature-controlled softmax:
\begin{align}
\tilde{s}_{2N-1}^i &= \frac{s_{2N-1}^i - \max(s_{2N-1}^i, s_{2N}^i)}{\tau}, \\
\tilde{s}_{2N}^i &= \frac{s_{2N}^i - \max(s_{2N-1}^i, s_{2N}^i)}{\tau},
\end{align}
\begin{equation}
\alpha_{2N-1}^i = \frac{e^{\tilde{s}_{2N-1}^i}}{e^{\tilde{s}_{2N-1}^i}+e^{\tilde{s}_{2N}^i}}, 
\quad
\alpha_{2N}^i = 1 - \alpha_{2N-1}^i .
\end{equation}

The final conditioning embedding is:
\begin{equation}
\mathbf{E}_i = \alpha_{2N-1}^i \mathbf{E}_{P_{2N-1}} + \alpha_{2N}^i \mathbf{E}_{P_{2N}} .
\end{equation}

To preserve spatial consistency, the cross-attention mask associated with the dominant prompt at timestep $i$ is selected and applied during denoising, ensuring that prompt emphasis affects only semantically relevant regions rather than the entire frame.

\vspace*{-0.5em}
\subsection{Time-Weighted Latent Space Blending (TWB)}
While DIPW improves intra-segment coherence, long-form continuity requires explicit cross-segment information transfer.
TWB initializes each segment using a temporally weighted summary of latent frames from the preceding segment.

Let $z_{(2N-3)(2N-2),i}$ denote the $i$-th latent frame of the previous segment, where the double index identifies the segment boundary rather than an individual frame.
We compute a blended initialization latent as:
\begin{equation}
\tilde{z}_{(2N-1)(2N),0} = \sum_{i=0}^{K-1} \tilde{w}_i \, z_{(2N-3)(2N-2),i},
\end{equation}
where $K$ is the number of tail frames considered. Weights follow an exponential decay:
\begin{equation}
w_i = \beta^{(K-i-1)}, \quad \tilde{w}_i = \frac{w_i}{\sum_{j=0}^{K-1} w_j},
\end{equation}
which emphasizes recent motion while suppressing outdated dynamics.
Exponential decay is chosen for its stability and monotonic temporal bias. The initial latent of the new segment is updated as:
\begin{align}
z_{(2N-1)(2N),0} \leftarrow\ & \gamma \cdot z_{(2N-3)(2N-2), T} + \gamma \cdot \tilde{z}_{(2N-1)(2N), 0} \notag \\
& + (1 - 2\gamma) \cdot z_{(2N-1)(2N), 0},
\end{align}
where $z_{(2N-3)(2N-2), T}$ denotes the final latent frame of the previous segment.
This update is applied once per segment transition and does not affect within-segment diffusion dynamics.

\vspace*{-0.5em}
\subsection{Semantic Action Representation (SAR)}
TWB alone may over-smooth transitions between semantically distinct actions.
To address this, SAR modulates blending strength based on action similarity. Each prompt $P_i$ is embedded using a pretrained text encoder:
\begin{equation}
\mathbf{a}_i = f_A(P_i),
\end{equation}
and action similarity is computed as:
\begin{equation}
S_A(P_{2N-1}, P_{2N}) = \frac{\mathbf{a}_{2N-1} \cdot \mathbf{a}_{2N}}{\|\mathbf{a}_{2N-1}\|\|\mathbf{a}_{2N}\|}.
\end{equation}

The effective blending coefficient is defined as:
\begin{equation}
\alpha' = \alpha \cdot (1 - S_A(P_{2N-1}, P_{2N})),
\end{equation}
where $\alpha$ is the base TWB blending factor. This formulation reduces temporal coupling for semantically similar actions while enabling stronger transitions for dissimilar actions.

\vspace*{-1em}
\section{Experiments and Results}

\begin{figure*}[htb!]
\begin{center}
\vspace*{-1em}
\includegraphics[width=0.9\linewidth]{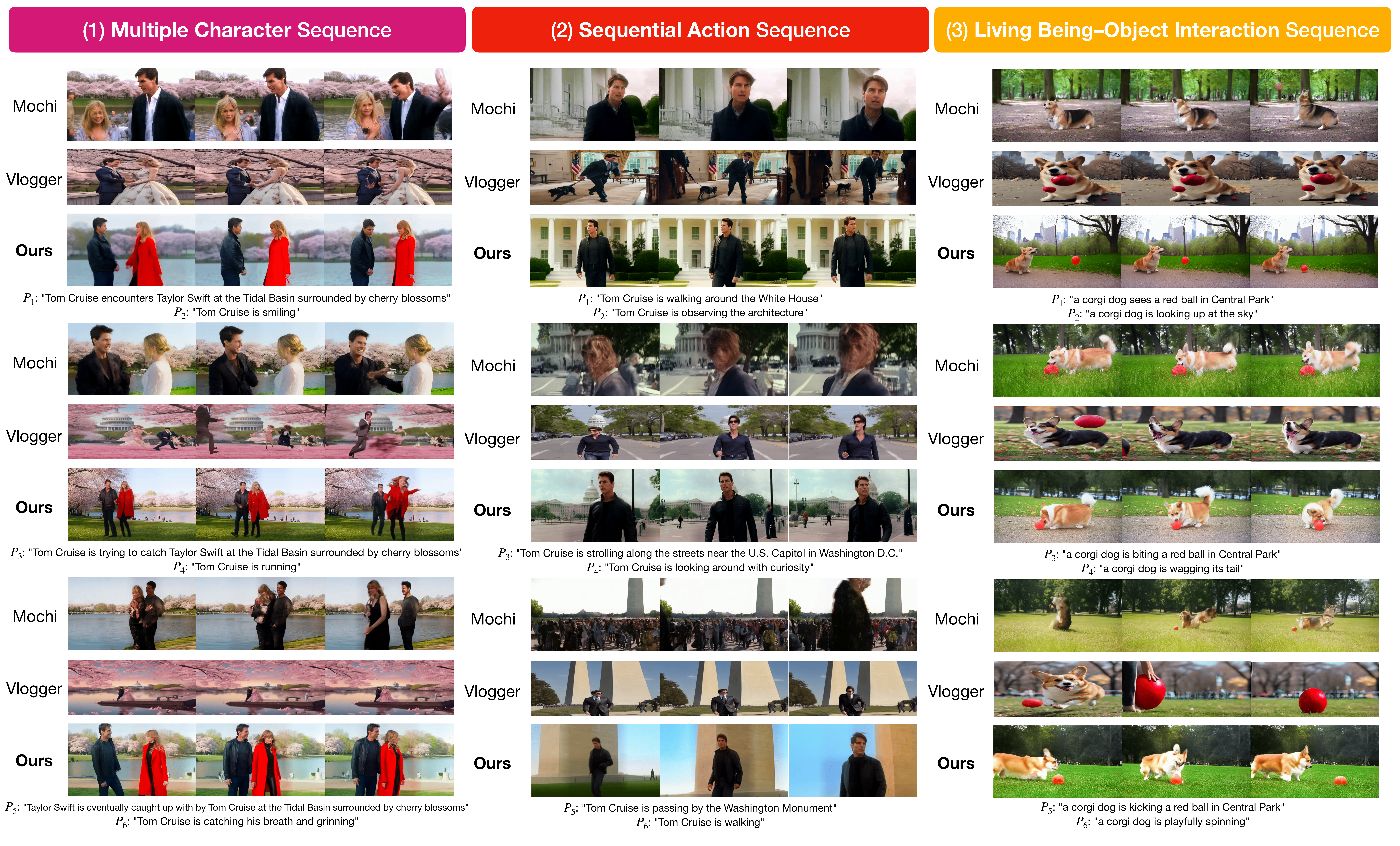}
\end{center}
\vspace*{-2em}
\caption{\textbf{Qualitative comparison across three sequence types:} 
(1) multiple character sequence, (2) sequential action sequence, and (3) living being–object interaction sequence. Our model consistently maintains character positioning, background consistency, motion continuity, and natural interactions across diverse scenarios. For additional video demonstrations, please refer to the supplementary material.}
\label{results:fig_all}
\vspace*{-1.25em}
\end{figure*}

\vspace*{-0.5em}
\subsection{Implementation Details} 
\vspace*{-0.5em}
We use Mochi-1~\cite{mochi} as the pretrained text-to-video backbone in an inference-only setting (no additional training). All videos are generated at $848\times480$ resolution with classifier-free guidance~\cite{ho2022classifier} scale 4.5 and 64 denoising steps. For prompt-level control, we apply our Dynamics-Informed Prompt Weighting (DIPW) at each diffusion timestep, where blending weights between the scene and action prompt embeddings are computed using (i) CLIP similarity, (ii) alignment with the previously applied embedding, and (iii) a diffusion-step-dependent prior, with $\lambda_1=\lambda_2=\lambda_3=1.0$ and temperature $\tau=0.5$. The dominant prompt selects the attention mask for conditioning. All experiments are conducted on a single NVIDIA H100 GPU (80GB). Additional implementation details are provided in Appendix~\ref{supp:implementation}.

\begin{table}[t]
\begin{small}
    \centering
    \vspace*{-1em}
    \resizebox{0.5\textwidth}{!}{%
    \begin{tabular}{lcccccc}
        \toprule
        \textbf{Method} & \textbf{CLIP-add} $\uparrow$ & \textbf{CLIP-combined} $\uparrow$ & \textbf{BLIP} $\uparrow$ & \textbf{DINO} $\uparrow$ & \textbf{LPIPS} ($V_{12} - V_{(2N-1)(2N)}$) $\downarrow$ \\
        \midrule
        Vlogger & 0.2889 & 0.2998 & 28.8509 & 0.9254 & \textbf{0.6193} \\
        \midrule 
        Mochi & 0.2940 & 0.3103 & \textbf{29.5948} & 0.9530 & 0.6811 \\
        \midrule
        \textbf{Ours (Full Model)} & \textbf{0.3055} & \textbf{0.3211} & 29.5117 & \textbf{0.9697} & 0.6733 \\
        \midrule
        Ours w/o DIPW (no prompt weigthing) & 0.2883 & 0.3013 & 28.9450 & 0.9418 & 0.6412 \\
        Ours w/o TWB (no temporal blending) & 0.2944 & 0.3030 & 29.1403 & 0.9655 & 0.6803 \\
        Ours w/o SAR (no semantic action rep) & 0.2926 & 0.3005 & 28.7187 & 0.9526 & \textbf{0.6396} \\
        \midrule
    \end{tabular}}
    \vspace*{-1em}
    \caption{\textbf{Quantitative evaluation of generated videos.} Higher CLIP and BLIP scores indicate better text alignment, while lower LPIPS values signify improved realism and story consistency. TWB indicates our temporally-aware latent space blending approach, SAR is the semantic action representation method and DIPW refers to the dynamics-informed prompt weighting (DIPW) based smoothing strategy. Our approach imparts significant benefits towards generating temporally consistent long-form videos, as also evidenced by the quantitative results.}
    \label{tab:quantitative_results}
    \vspace*{-1.25em}
\end{small}
\end{table}

\vspace*{-0.5em} 
\subsection{Dataset} 
\vspace*{-0.5em} 
We construct a prompt-pair storytelling dataset tailored to long-form text-driven video generation, where each story is represented as a sequence of paired prompts (scene description, action command). The dataset spans eight locations (seven with multi-character interactions), and each story contains 12--13 sequential scenes. We include two character sets (celebrity and animal) with four distinct character configurations across locations and scenes, resulting in 404 prompts in total. Further dataset details are provided in Appendix~\ref{supp:dataset}. \textbf{Baselines.} Since prior work does not directly support story-driven generation solely from text prompt pairs, we compare against the Mochi backbone with identical generation hyperparameters, and adapt Vlogger~\cite{zhuang2024vlogger} to our setting.

\vspace*{-0.5em}
\subsection{Qualitative Results} 
\vspace*{-0.5em}
Fig.~\ref{results:fig_all} compares long-form excerpts from stories consisting of 12--13 segments generated from discrete prompt pairs. Mochi typically produces independent short clips with abrupt transitions, while Vlogger captures motion patterns but often fails to maintain long-term consistency across segments. In contrast, our method produces smoother segment-to-segment transitions while preserving object structure, scene composition, and character appearance. Additional qualitative results are provided in Appendix~\ref{supp:qualitative}.

\vspace*{-1em} 
\subsection{Quantitative Evaluation} 
\vspace*{-0.5em}
\label{sec:quantitative_evaluation} 
We evaluate text alignment, coherence, and transition smoothness using CLIP (CLIP-combined and CLIP-add), BLIP, DINO, and LPIPS. Metrics are computed on a validation split comprising 8 stories (25\% of 32 stories), following prior evaluation practices~\cite{lin2023videodirectorgpt,wang2024dreamrunner}. As shown in Table~\ref{tab:quantitative_results}, our method achieves higher text alignment (CLIP/BLIP), stronger semantic coherence (DINO), and smoother transitions (lower LPIPS) than Mochi and Vlogger. While removing SAR can slightly reduce LPIPS, SAR explicitly encourages semantically meaningful action changes, which may increase perceptual change while improving narrative faithfulness. Detailed evaluation protocol and additional analyses are provided in Appendix~\ref{supp:detailed_quantitative_evaluation}.

\vspace*{-1em} 
\subsection{Ablation Study} 
\vspace*{-0.5em}
We conduct an ablation study to evaluate the contributions of Time-Weighted Blending (TWB), Dynamics-Informed Prompt Weighting (DIPW), and Semantic Action Representation (SAR) to long-form video generation. Qualitative results in Fig.~\ref{results:ablation} (Appendix) and quantitative comparisons in Table~\ref{tab:quantitative_results} show that the full model achieves the most coherent storytelling, maintaining scene consistency, smooth motion transitions, and logical action progression. Removing any component degrades temporal continuity, prompt adherence, or motion coherence, highlighting the complementary roles of TWB, DIPW, and SAR. Additional qualitative comparisons and analysis are provided in Appendix~\ref{supp:ablation}.

\vspace*{-1em} 
\section{Conclusion} 
\vspace*{-1em} 
We presented a training-free framework for long-form text-driven video generation that integrates dynamics-informed prompt weighting, time-weighted latent blending, and semantic action modeling. By explicitly separating scene context and action dynamics and enforcing temporally-aware control during diffusion, our method generates coherent multi-segment video narratives from discrete prompts. Extensive experiments demonstrate consistent improvements over strong baselines, effectively bridging the gap between short-form video synthesis and long-form storytelling. 

\noindent {\bf Limitations and Future Work: } 
Our approach may exhibit minor artifacts at segment boundaries in highly complex motion scenarios and assumes sufficient semantic continuity in the input prompts. Future work may  extend the framework to handle more dynamic transitions and interactive video generation settings with real-time user feedback.

\vspace*{0.1em}
\noindent
{\bf Acknowledgement: } This work is supported in part by Dr. Barry Mersky and Capital One E-Nnovate Endowed Professorships and University of Maryland Distinguished University
Professorship. 

\bibliographystyle{IEEEbib}
\bibliography{main}

\newpage
\appendix
\onecolumn

\section{Supplementary Materials}

\subsection{Ethics Statement}

\begin{tcolorbox}[breakable, title=Ethics Statement]
This work uses fully synthetic stories generated by an AI model solely for the purpose of qualitative and quantitative evaluation of generative storytelling systems. Some prompts and generated outputs include the names of well-known public figures (e.g., actors, etc) as narrative anchors to facilitate consistent character grounding in multimodal generation. These names are used strictly as fictional narrative identifiers and do not refer to, depict, or make claims about the real individuals, their lives, actions, beliefs, or historical events.

All stories presented in this work are entirely fictional and constructed in imaginary settings. They are not intended to resemble real events, nor to portray, endorse, criticize, or misrepresent any actual person, organization, culture, or political position. The generated content does not aim to reflect factual information and should not be interpreted as commentary on real individuals or used for any decision-making purpose.

The use of public-figure names is limited to controlled, non-defamatory, non-realistic scenarios and serves only as a methodological choice for evaluating character consistency and narrative coherence in generative models. No personal data is used, and no attempt is made to simulate real-world behaviors, speech, or private attributes of the referenced individuals.
\end{tcolorbox}

\subsection{Implementation Details}
\label{supp:implementation}
We use the Mochi-1~\cite{mochi} backbone model as a T2V(Text-to-Video) model in all our experiments. The model operates in an inference-only setting, without any additional training. The video resolution is set to $848 \times 480$, and we use a classifier-free guidance scale~\cite{ho2022classifier} of 4.5. The inference process involves 64 denoising steps per generation to ensure high-quality outputs. To enable prompt-level control during generation, we adopt the proposed Dynamics-Informed Prompt Weighting (DIPW) strategy. At each timestep, DIPW computes the blending weights of two prompt embeddings based on three components: (i) the CLIP similarity between the current frame and each prompt, (ii) temporal alignment with the previously applied embedding, and (iii) a diffusion-step-dependent prior. These components are equally weighted using fixed hyperparameters $\lambda_1 = \lambda_2 = \lambda_3 = 1.0$. The prompt with the higher weight determines the attention mask to maintain semantic focus and structural coherence. To avoid ambiguous weighting and encourage decisive transitions, the final weights are softly normalized such that one prompt serves as the dominant conditioning source while the other provides minimal but stabilizing influence. This weighting behavior was found to support both semantic alignment and motion continuity across long-form video segments. In our experiments, we use a fixed temperature of $\tau=0.5$ to balance selectivity and stability during prompt interpolation. All experiments (see Appendix document) are conducted on a single H100 GPU (80GB).

\subsection{Computation Time and Memory Consumption}
\label{supp:quantitative}

\begin{table}[htb!]
    \small
    \centering
    \resizebox{0.8\textwidth}{!}{
    \begin{tabular}{l|l|c|c}
    \toprule
    \textbf{Model} & \textbf{Step} & \textbf{Memory Consumption} & \textbf{Computation Time} \\
    \midrule
    Full Model & Inference & 31.2276 GB / 80.0 GB (29,781 MiB) & 880 sec \\
    \midrule
    w/o DIPW & Inference & 31.2108 GB / 80.0 GB (29,765 MiB) & 357 sec \\
    \midrule
    w/o TWB & Inference & 31.2234 GB / 80.0 GB (29,777 MiB) & 917 sec \\
    \midrule
    w/o SAR & Inference & 30.5775 GB / 80.0 GB (29,161 MiB) & 1,205 sec \\
    \midrule
    w/o ALL (Mochi) & Inference & 25.2801 GB / 80.0 GB (24,109 MiB) & 126 sec \\
    \bottomrule
    \end{tabular}
    }
    \caption{\textbf{Step-wise Memory Consumption and Computation Time Analysis}. This table presents a detailed comparison of memory consumption and computation time for our model and its ablated variants during inference, evaluated on an NVIDIA H100 GPU (80GB). The ``Full Model'' includes all proposed components, achieving a balanced trade-off between efficiency and quality. Removing TWB slightly increases inference time, while excluding DIPW significantly reduces computation time. In contrast, removing SAR leads to a substantial increase in computation time, highlighting its role in optimization. The baseline ``w/o ALL (Mochi)'' configuration has the lowest memory and fastest inference time but lacks all of the benefits we mentioned in this paper.}
    \label{computation-comparison}
\end{table}

Table~\ref{computation-comparison} presents a detailed comparison of memory consumption and computation time across different configurations of our model, evaluated using an NVIDIA H100 GPU (80GB). The analysis highlights the impact of various components on inference efficiency, demonstrating how each contributes to overall computational requirements.

Our full model achieves high-quality synthesis with a memory consumption of 31.21 GB and an inference time of 869 seconds. The result demonstrates that while individual components contribute to different aspects of computational performance, our full model strikes a balance between efficiency and performance. The ablation study confirms that SAR plays a crucial role in speeding up inference, while TWB and DIPW contribute to reducing overall computational time. Notably, the baseline model (Mochi) is the most lightweight but lacks the high-fidelity outputs achieved by the full model.

Overall, our approach effectively balances computational cost and memory efficiency, making it suitable for real-world applications where both scalability and high-quality synthesis are critical.

\subsection{Detailed Qualitative Results}
\label{supp:qualitative}
Left one in Figure~\ref{results:fig_all} presents a qualitative comparison of a multi-character interaction sequence where Tom Cruise meets Taylor Swift at the Tidal Basin. Our model effectively preserves relative positioning and interactions between the two characters, while the baseline methods distort spatial relationships or fail to maintain interaction consistency. The background remains visually stable in our results, while competing methods introduce inconsistencies in the cherry blossom setting. Our model also captures emotional progression, with Tom Cruise transitioning from smiling to running and catching his breath, while other methods often fail to maintain facial expression consistency across frames. Furthermore, logical action continuity is evident in our results, as Tom Cruise’s movement smoothly follows the pursuit-and-capture narrative, whereas baseline models frequently introduce abrupt or disjointed transitions. Also, the bottom one illustrates a sequential movement sequence where Tom Cruise walks through Washington, D.C., passing key landmarks such as the White House and the Washington Monument. Our model ensures smooth temporal continuity, where each motion naturally leads into the next, while some baselines generate erratic or inconsistent movements. Unlike other methods that frequently reset poses abruptly between frames, our approach maintains the logical impact of prior movements on subsequent actions. Scene awareness is also preserved, ensuring architectural landmarks remain stable, while competing methods often introduce background distortions or misplaced elements. The right one demonstrates an interaction between a corgi and a red ball in Central Park. Our model ensures object permanence, keeping the ball consistently positioned and preventing unnatural displacement, while other methods often fail to track the object properly. The action sequence follows a natural progression from the corgi seeing the ball, biting it, and then kicking it, whereas some baselines introduce inconsistencies by skipping intermediate actions or generating illogical motion transitions. Our model also captures realistic canine behavior, such as tail wagging and playful spinning, while other methods often produce rigid or unnatural movements.

\subsection{Ablation Study}
\label{supp:ablation}

Without Time-weighted Blending (TWB), the generated frames lack temporal consistency, resulting in abrupt scene transitions where Tom Cruise and Taylor Swift appear as different identities across frames. Additionally, there is no meaningful interaction between them, making the sequence feel disconnected. The full model utilizes TWB to enforce bidirectional constraints, preserving spatial and temporal continuity across video segments. Without Dynamics-Informed Prompt Weighting (DIPW), the model fails to integrate Prompt 2 into the scene, leading to incomplete or inaccurate action sequences. The absence of DIPW prevents the model from smoothly interpolating between different prompt levels, causing a loss of intended action details and reducing narrative control. Our approach leverages DIPW to guide structured prompt blending, ensuring accurate action progression aligned with the evolving scene. Without Semantic Action Representation (SAR), the continuity of motion and action sequences deteriorates. The absence of SAR leads to disjointed character actions, where movements do not logically connect across frames, disrupting motion coherence. The full model incorporates SAR to encode high-level action semantics, ensuring that character behaviors evolve naturally and respond dynamically to preceding movements. These results highlight the necessity of each component: TWB maintains scene and identity consistency, DIPW enables structured prompt-driven action transitions, and SAR ensures smooth motion continuity and logical action sequences. The full integration of these modules allows for semantically aligned, visually coherent, and perceptually smooth long-form video generation. 

Table~\ref{tab:quantitative_results} shows our evaluation results for ablation study. The full model (Ours) achieves the best overall performance, demonstrating that each module plays a crucial role in generating coherent long-form videos. Although LPIPS is lower when SAR is removed (0.6396), this does not indicate better video quality. Instead, it reflects reduced motion complexity and less dynamic transitions, as SAR enhances character interactions and logical action continuity at the cost of slightly increased perceptual differences. The drop in CLIP-add (0.2926) and CLIP-combined (0.3005) without SAR further confirms that it is essential for maintaining text-video alignment. Similarly, removing DIPW leads to weaker prompt-based scene transitions, while TWB removal results in the highest LPIPS (0.6887), indicating degraded temporal smoothness. These results highlight that SAR, DIPW, and TWB must be combined to ensure text-aligned, semantically structured, and perceptually coherent video generation. The method for quantitative evaluation follows Section~\ref{sec:quantitative_evaluation}.

\subsection{Detailed Quantitative Evaluation}
\label{supp:detailed_quantitative_evaluation}
Focusing on temporal coherence, spatial consistency, and smooth transitions across video segments while preserving the semantic fidelity of input text prompts, Figures~\ref{results:fig_all} illustrate results from our model along with comparisons. All frame comparisons for the videos are provided based on the original resolution. Each sequence is part of a larger generated story consisting of 12 to 13 video segments, from which we have selected key excerpts for comparison. Each sequence consists of multiple video segments \(V_{12}, V_{34}, V_{56}, \dots\) generated from discrete prompts \( (P_1, P_2), (P_3, P_4), (P_5, P_6), \dots \). Comparisons reveal that Mochi generates independent short clips without explicit transitions and has abrupt scene changes. Vlogger captures motion patterns but struggles with maintaining long-term consistency and coherent transitions. These results highlight the key advantages of our approach. Our method maintains temporal consistency by leveraging time-weighted blending, which ensures motion continuity, prevents sudden appearance changes, and preserves spatial stability by maintaining object structure, scene composition, and character placement. Semantic coherence is achieved through prompt mixing techniques that enhance alignment between generated frames and textual descriptions. Additionally, spatial-attention refinement (SAR) plays a crucial role in enhancing local consistency by refining fine-grained details, reducing spatial artifacts, and ensuring that character features, backgrounds, and object interactions remain visually stable across frames. Overall, our approach significantly outperforms existing baselines by providing a structured, seamless, and coherent long-form video synthesis framework.

We evaluate the quality of generated videos using multiple metrics to measure text alignment, visual fidelity, and story consistency. CLIP Score is employed in two variants to assess text alignment: CLIP-combined, which measures overall alignment by comparing the generated frame against a composite representation of all individual text prompts, and CLIP-add, which computes the average CLIP score across individual text prompts to capture alignment with specific concepts. Higher CLIP scores indicate better text-video consistency. Additionally, the BLIP Score evaluates text-level alignment by comparing the generated frame against a combined representation of all text prompts, ensuring that the generated content aligns well with the intended textual descriptions. For visual and structural consistency, we employ DINO to measure frame-level semantic similarity, and the Learned Perceptual Image Patch Similarity (LPIPS) metric to evaluate perceptual continuity across segments. We compute the average LPIPS scores across all consecutive video segments, specifically from $V_{12}$ to $V_{(2N-1)(2N)}$, to quantify the extent to which our method maintains temporal and spatial continuity throughout the generated story. Lower LPIPS values indicate smoother transitions between segments. Considering the evaluation methodology of VideoDirectorGPT~\cite{lin2023videodirectorgpt} and the dataset scale used in the DreamRunner~\cite{wang2024dreamrunner} (DreamStorySet, which consists of 13 stories with 5 to 8 scenes each) study, we use 25\% of the overall data (8 stories from a total of 32 stories) from our dataset as the validation set to compute the quantitative metrics. Table~\ref{tab:quantitative_results} presents the quantitative comparison of our method against baseline approaches, including Mochi and Vlogger. Our approach achieves superior text alignment (higher CLIP scores), stronger frame-level semantic coherence (higher DINO), and enhanced story consistency (lower LPIPS), demonstrating its effectiveness in generating long-form coherent video sequences. Notably, removing SAR yields a slightly lower LPIPS score, but this does not indicate better temporal consistency. SAR explicitly models semantically meaningful action transitions, which introduces purposeful perceptual changes between segments. As a result, LPIPS marginally increases, but the transitions become more narratively coherent and visually faithful to the intended actions—reflecting improved story quality rather than degradation.

Table~\ref{tab:quantitative_results_std} reports the same quantitative evaluation as Table~\ref{tab:quantitative_results}, but includes standard deviation values (mean ± std) computed across 8 validation stories. The inclusion of standard deviations provides insight into the stability and consistency of each method across diverse prompts and scenes. Our full model not only achieves the best average performance across all metrics—including CLIP-add, CLIP-combined, BLIP, DINO, and LPIPS—but also shows stable results with relatively low variance. Each ablation variant (w/o DIPW, TWB, SAR) demonstrates noticeable drops in performance or increased variability, highlighting the contribution of each component to the overall video generation quality.

\label{supp:Qualitative}

\begin{table}[t]
\centering
\small
\setlength{\tabcolsep}{4pt}
\resizebox{\textwidth}{!}{%
\begin{tabular}{lccccc}
\toprule
\textbf{Method} & \textbf{CLIP-add} $\uparrow$ & \textbf{CLIP-combined} $\uparrow$ & \textbf{BLIP} $\uparrow$ & \textbf{DINO} $\uparrow$ & \textbf{LPIPS} ($V_{12} - V_{(2N-1)(2N)}$) $\downarrow$ \\
\midrule
Vlogger & 0.2889 ± 0.0133 & 0.2998 ± 0.0122 & 28.8509 ± 1.2651 & 0.9254 ± 0.0188 & \textbf{0.6193} ± 0.0214 \\
Mochi & 0.2940 ± 0.0139 & 0.3103 ± 0.0180 & \textbf{29.5948} ± 1.8717 & 0.9530 ± 0.0098 & 0.6811 ± 0.0386 \\
\textbf{Ours (Full Model)} & \textbf{0.3055} ± 0.0095 & \textbf{0.3211} ± 0.0122 & 29.5117 ± 1.6253 & \textbf{0.9697} ± 0.0056 & 0.6733 ± 0.0501 \\
Ours w/o DIPW & 0.2883 ± 0.0214 & 0.3013 ± 0.0213 & 28.9450 ± 1.7390 & 0.9418 ± 0.0131 & 0.6412 ± 0.0224 \\
Ours w/o TWB & 0.2944 ± 0.0152 & 0.3030 ± 0.0157 & 29.1403 ± 1.1490 & 0.9655 ± 0.0093 & 0.6803 ± 0.0440 \\
Ours w/o SAR & 0.2926 ± 0.0107 & 0.3005 ± 0.0124 & 28.7187 ± 0.8393 & 0.9526 ± 0.0150 & \textbf{0.6396} ± 0.0324 \\
\bottomrule
\end{tabular}}
\caption{\textbf{Quantitative evaluation of generated videos (mean ± std).} Our method shows the best or comparable performance across multiple metrics. Lower LPIPS indicates better realism and temporal consistency. Standard deviations are omitted for brevity.}
\label{tab:quantitative_results_std}
\end{table}

\subsection{Dataset Plot and Character}
\label{supp:dataset}
Coherent \emph{story generation} requires not only visual consistency within a scene but also continuity across actions that drive the narrative forward. Unlike single-prompt video generation, where one caption is sufficient, storytelling inherently demands the interplay between \emph{scene descriptions} (to ground the environment and context) and \emph{action commands} (to advance the plot). However, no existing dataset directly supports this structured prompt-pair setting for long-form video storytelling. Inspired by Set 4 of ~\cite{kothandaraman2024prompt} for prompt image-based mixing, we construct a diverse data set for our task that includes both animate objects (e.g. humans, animals) and inanimate objects (e.g. vehicles, sports equipment). The dataset spans eight locations, with seven featuring multi-character interactions. Each story consists of 12-13 sequential scenes, and we introduce two character sets: a celebrity set and an animal set. With four distinct character configurations across all locations and scenes, the dataset comprises 404 prompts, providing a structured and diverse benchmark for evaluating story generation models. A detailed description of the dataset can be found in Appendix~\ref{supp:dataset}. \textbf{Baselines.} In the existing literature, no method can generate story-driven videos solely from text, making ours the first to do so. To comprehensively evaluate the effectiveness of our method, we compare it with the corresponding Mochi backbone. We use the same generation hyperparameters to keep baseline comparisons consistent. We also compare our method with a SOTA story generation method, Vlogger~\cite{zhuang2024vlogger}, adapted for our problem. 

Our dataset is designed to comprehensively represent both animate (e.g., humans, animals) and inanimate objects (e.g., balls, buses, cars, boats, airplanes) to ensure diversity in story generation. Each story plot includes at least one visually distinguishable action performed by an entity, such as throwing a ball, boarding a vehicle, running, pressing a button, or walking, to enhance dynamic storytelling. 

The dataset covers eight diverse locations: New York City, Washington D.C., Paris, London, Los Angeles, San Francisco, Chicago, and Las Vegas. Among these, seven locations (excluding New York City) feature at least two characters (or two animals) per plot, introducing interactions and multi-character dynamics. Each story plot consists of 12 to 13 sequential scenes, with Chicago, Las Vegas, and New York City including 12 scenes per story, while all other locations include 13 scenes per story. To ensure diversity in character representation, we introduce two distinct character sets: a celebrity set (Tom Cruise \& Taylor Swift, Elon Musk \& Angelina Jolie) and an animal set (Corgi Dog \& Siamese Cat, Panda \& Fox). Given eight different locations, 12-13 scenes per story, and four distinct character settings, the dataset consists of a total of 404 prompts. ($12 \times 3 + 13 \times 5 = 101$, $101 \times 4 = 404$) 

Prompt 1 provides a broad scene description that establishes the setting and context, while Prompt 2 introduces specific actions performed by the character within that scene. For example, in a New York City subway scenario, Prompt 1 may be “Tom Cruise is inside of the subway train,” setting up the environment, whereas Prompt 2 specifies an action such as “Tom Cruise is sitting.” This structure enables fine-grained control over character movement and interactions while maintaining coherence in scene transitions.

This structured approach allows for a broad range of environments, interactions, and character-driven narratives, making it well-suited for evaluating story generation models. Actions that humans can perform but animals cannot may be adapted accordingly. For example, since a dog cannot pick up and throw a ball with both hands, such an action is replaced with the dog kicking the ball instead.

To be specific, in a scene where the character arrives at Central Park, Prompt 1 describes, “Tom Cruise sees a red ball in Central Park,” providing situational context, while Prompt 2 refines the action in detail with, “Tom Cruise is picking up a red ball in Central Park” and later “Tom Cruise is throwing a red ball in Central Park.” Similarly, in an animal-based variation, Prompt 1 states, “A corgi dog sees a red ball in Central Park,” while Prompt 2 adapts the action appropriately, such as “A corgi dog is biting a red ball in Central Park” and “A corgi dog is kicking a red ball in Central Park.” 

This approach ensures that actions are naturally adapted for different entities, particularly when an action performed by a human (e.g., gripping and throwing a ball) must be substituted with a more plausible behavior for an animal (e.g., biting or kicking the ball). The dataset maintains consistency in narrative progression while allowing for variations based on both character type and setting. Some examples of dataset plots can be found below.

\begin{lstlisting}
prompt_nyc = [
    "Tom Cruise is inside of the subway train", "Tom Cruise is sitting",
    "Tom Cruise is looking out the subway window", "Tom Cruise now stands out",
    "Tom Cruise is getting off the NYC subway train", "Tom Cruise is walking",
    "Tom Cruise is walking up the subway exit stairs", "Tom Cruise is looking around",
    "Tom Cruise is looking at the streets of Times Square, NYC", "Tom Cruise is tilting his head curiously",
    "Tom Cruise is walking on the streets of Times Square, NYC", "Tom Cruise is walking",
    "Tom Cruise is waiting for a bus at the Times Square bus stop in NYC", "Tom Cruise is standing",
    "Tom Cruise is getting on a bus at the Times Square bus stop", "Tom Cruise is walking",
    "Tom Cruise is looking out the bus window at the city view", "Tom Cruise is sitting",
    "Tom Cruise has arrived at Central Park", "Tom Cruise is strolling",
    "Tom Cruise sees a red ball in Central Park", "Tom Cruise is observing it curiously",
    "Tom Cruise is picking up a red ball in Central Park", "Tom Cruise is gripping it firmly",
    "Tom Cruise is throwing a red ball in Central Park", "Tom Cruise is watching its trajectory"
]
prompt_nyc_corgi = [
    "a corgi dog is inside of the subway train", "a corgi dog is sitting",
    "a corgi dog is looking out the subway window", "a corgi dog now stands out",
    "a corgi dog is getting off the subway train", "a corgi dog is walking",
    "a corgi dog is looking at the streets of Times Square, NYC", "a corgi dog is tilting its head curiously",
    "a corgi dog is walking on the streets of Times Square, NYC", "a corgi dog is wagging its tail",
    "a corgi dog is waiting for a bus at the Times Square bus stop in NYC", "a corgi dog is standing",
    "a corgi dog is getting on a bus at the Times Square bus stop", "a corgi dog is walking",
    "a corgi dog is looking out the bus window at the city view", "a corgi dog is sitting",
    "a corgi dog has arrived at Central Park", "a corgi dog is sniffing the ground",
    "a corgi dog sees a red ball in Central Park", "a corgi dog is looking up at the sky",
    "a corgi dog is biting a red ball in Central Park", "a corgi dog is wagging its tail",
    "a corgi dog is kicking a red ball in Central Park", "a corgi dog is playfully spinning"
]
prompt_dc = [
    "Tom Cruise is walking around the White House", "Tom Cruise is observing the architecture",
    "Tom Cruise is strolling along the streets near the U.S. Capitol in Washington D.C.", "Tom Cruise is looking around with curiosity",
    "Tom Cruise is passing by the Washington Monument", "Tom Cruise is walking",
    "Tom Cruise is walking along the Tidal Basin surrounded by cherry blossoms", "Tom Cruise is taking a leisurely stroll",
    "Tom Cruise stops for a moment at the Tidal Basin surrounded by cherry blossoms", "Tom Cruise is sitting",
    "Tom Cruise is jogging along the Tidal Basin surrounded by cherry blossoms", "Tom Cruise is enjoying the fresh air",
    "Tom Cruise encounters Taylor Swift at the Tidal Basin surrounded by cherry blossoms", "Tom Cruise is smiling",
    "Tom Cruise is trying to catch Taylor Swift at the Tidal Basin surrounded by cherry blossoms", "Tom Cruise is running",
    "Taylor Swift is eventually caught up with by Tom Cruise at the Tidal Basin surrounded by cherry blossoms", "Tom Cruise is catching his breath and grinning",
    "Tom Cruise and Taylor Swift are enjoying the cherry blossoms together", "Tom Cruise and Taylor Swift are sitting side by side",
    "Tom Cruise and Taylor Swift are admiring the cherry blossoms at the Tidal Basin", "Tom Cruise and Taylor Swift are lying on the grass looking up at the sky",
    "Tom Cruise and Taylor Swift are lying on the grass at the Tidal Basin surrounded by cherry blossoms, slowly closing their eyes", "Tom Cruise and Taylor Swift are resting peacefully",
    "Tom Cruise and Taylor Swift have fallen asleep at the Tidal Basin surrounded by cherry blossoms", "Tom Cruise and Taylor Swift are peacefully dozing off"
]
prompt_dc_corgi = [
    "a corgi dog is walking around the White House", "a corgi dog is sniffing the ground",
    "a corgi dog is strolling along the streets near the U.S. Capitol in Washington D.C.", "a corgi dog is walking",
    "a corgi dog is passing by the Washington Monument", "a corgi dog is looking around",
    "a corgi dog is walking along the Tidal Basin surrounded by cherry blossoms", "a corgi dog is taking a leisurely stroll",
    "a corgi dog stops for a moment at the Tidal Basin surrounded by cherry blossoms", "a corgi dog is sitting",
    "a corgi dog is running along the Tidal Basin surrounded by cherry blossoms", "a corgi dog is excitedly running",
    "a corgi dog encounters a siamese cat at the Tidal Basin surrounded by cherry blossoms", "a corgi dog is tilting its head curiously",
    "a corgi dog is chasing a fleeing siamese cat with at the Tidal Basin surrounded by cherry blossoms", "a corgi dog and a siamese cat are running",
    "a siamese cat is finally caught by a corgi dog at the Tidal Basin surrounded by cherry blossoms", "a corgi dog is gently wagging its tail",
    "a corgi dog and a siamese cat are sniffing the scent of cherry blossoms together at the Tidal Basin", "a corgi dog and a siamese cat are sitting",
    "a corgi dog and a siamese cat are enjoying the cherry blossoms at the Tidal Basin", "a corgi dog and a siamese cat are lying down",
    "a corgi dog and a siamese cat are lying down at the Tidal Basin surrounded by cherry blossoms, slowly closing their eyes", "a corgi dog and a siamese cat are lying down",
    "a corgi dog and a siamese cat have fallen asleep at the Tidal Basin surrounded by cherry blossoms", "a corgi dog and a siamese cat are peacefully sleeping"
]
(...)
\end{lstlisting}

\begin{figure}[htb!]
\begin{center}
\includegraphics[width=0.5\linewidth]{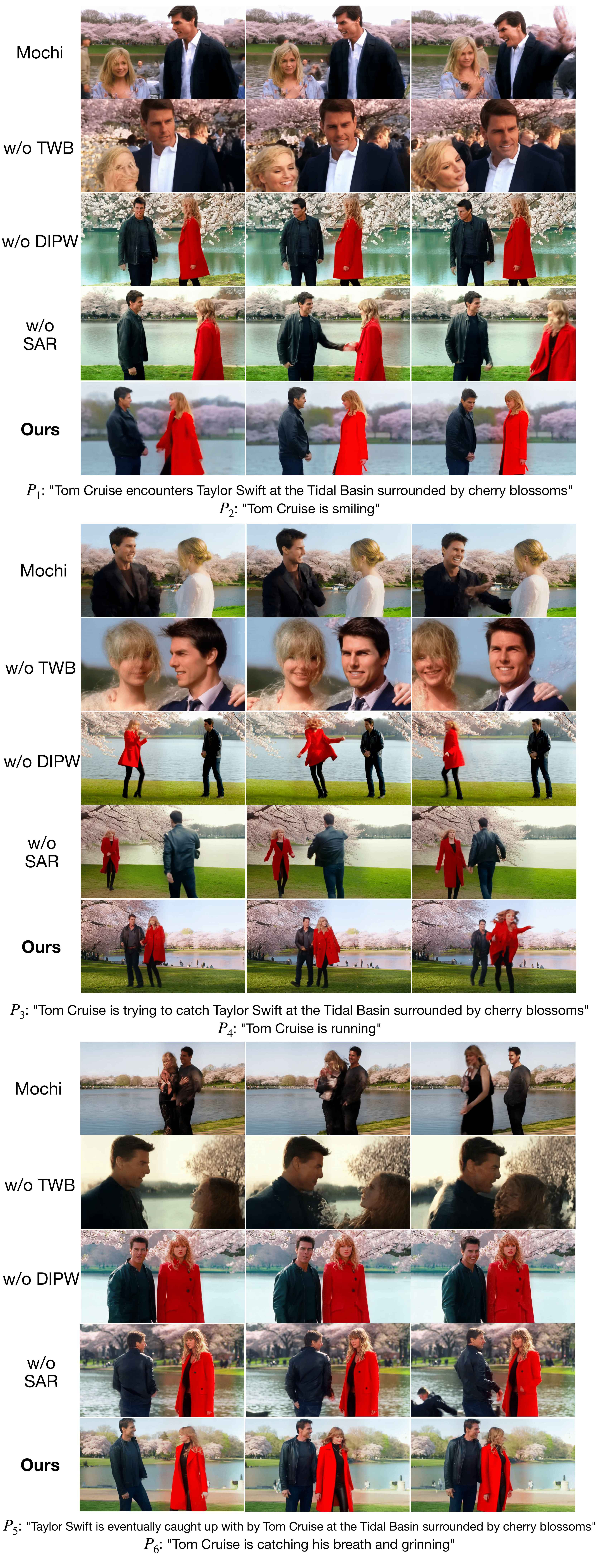}
\end{center}
\vspace*{-1.5em}
    \caption{\textbf{Ablation Study} Each row shows a different setting: Full Model, w/o TWB, w/o DIPW, w/o SAR, and a baseline (Mochi). The full model produces the most coherent motion and character interactions. }
\label{results:ablation}
\vspace*{-2em}
\end{figure}

\end{document}